\documentclass{article}
\usepackage{spconf,amsmath,graphicx}

\usepackage{fancyhdr}
\usepackage{enumitem}
\usepackage{booktabs}
\setlist{nosep, leftmargin=14pt}

\usepackage{mwe} 

\title{Distilling Expert Surgical Knowledge: How to train local surgical VLMs for anatomy explanation in Complete  Mesocolic Excision}
\name{Lennart Maack$^{1}$, Julia-Kristin Graß$^{2}$, Lisa-Marie Toscha$^{2}$, Nathaniel Melling$^{2}$, Alexander Schlaefer$^{1}$ \thanks{mail: lennart.maack@tuhh.de}}
\address{$^{1}$ Institute of Medical Technology and Intelligent Systems, \\ Hamburg University of Technology, Hamburg, Germany \\
$^{2}$ Department of General, Visceral and Thoracic Surgery, \\ University Medical Center Hamburg-Eppendorf, Hamburg, Germany}

\begin{document}
%
\maketitle

\thispagestyle{fancy}

\begin{abstract}

    Recently, Vision Large Language Models (VLMs) have demonstrated high potential in computer-aided diagnosis and decision-support.
    However, current VLMs show deficits in domain specific  surgical scene understanding, such as identifying and explaining anatomical landmarks during Complete Mesocolic Excision.
    Additionally, there is a need for locally deployable models to avoid patient data leakage to large VLMs, hosted outside the clinic. 
    We propose a privacy-preserving framework to distill knowledge from large, general-purpose LLMs into an efficient, local VLM.
    We generate an expert-supervised dataset by prompting a teacher LLM without sensitive images, using only textual context and binary segmentation masks for spatial information.
    This dataset is used for Supervised Fine-Tuning (SFT) and subsequent Direct Preference Optimization (DPO) of the locally deployable VLM.
    Our evaluation confirms that finetuning VLMs with our generated datasets increases surgical domain knowledge compared to its base VLM by a large margin.
    Overall, this work validates a data-efficient and privacy-conforming way to train a surgical domain optimized, locally deployable VLM for surgical scene understanding.
\end{abstract}

\section{Introduction}
\begin{figure}[t]
\centering
\centerline{\includegraphics[width=0.95\columnwidth]{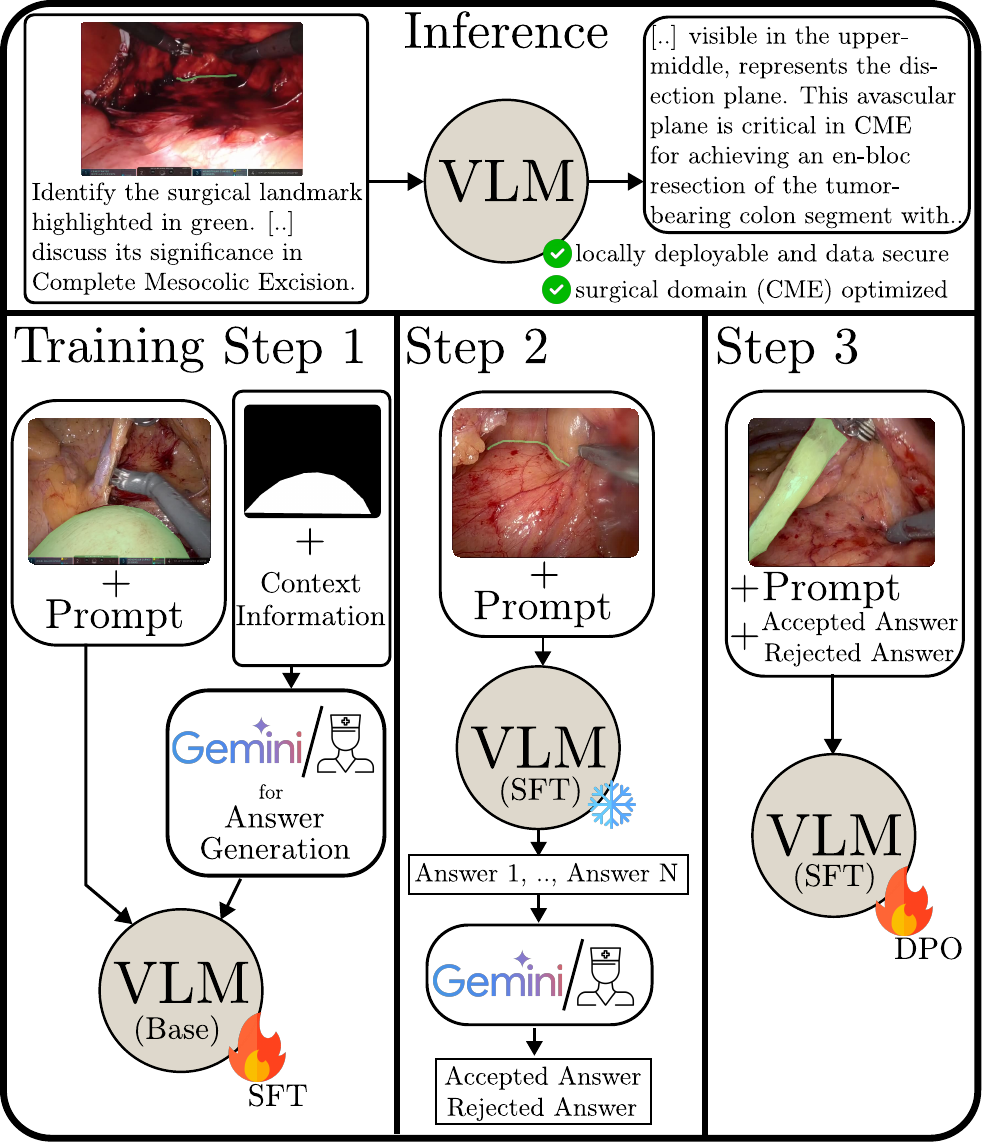}}
\caption{Overview of our proposed training framework for training a surgical domain optimized, locally deployable Vision Language Model (VLM).}\medskip
\label{fig:Figure_1}
\end{figure}
Robot-assisted Complete Mesocolic Excision (CME) has gained prominence due to its oncological benefits \cite{spinoglio2018robotic}.
A critical and challenging step in this surgery is the meticulous dissection along the mesofacial interface, where the colorectal specimen is mobilized from the retroperitoneum.
Maintaining the intactness of the mesocolic fascial envelope is of importance for high-quality oncological resection \cite{dimitriou2015complete}.
However, this phase presents difficulties, as correctly identifying and interpreting critical intraoperative anatomical landmarks, such as the correct plane of dissection are hard, especially for less experienced surgeons. \\
Computer-assisted surgery (CAS) aims to support surgeons during surgery or surgical training by providing consistent, real-time support \cite{maier2017surgical}.
Despite impressive results in the development of algorithms for individual surgical tasks such as anatomy segmentation or workflow recognition in the field of CAS \cite{maack2024efficient, twinanda2016endonet}, these task-specific solutions have limited potential in providing context-related explanations.
More comprehensive assistance in the form of visual support, combined with explanation in natural language can lead to a better overall understanding of the surgical scene \cite{seenivasan2022surgical}.\\
Vision Language Models (VLMs) integrate the natural language capabilities of Large Language Models with the visual perception of vision encoders and offer a solution to the current limitations in CAS by serving as versatile surgical understanding systems. \\
However, the use of general-purpose models for surgical applications has several disadvantages.
The use of cloud-based models such as GPT-5 or Gemini requires the transfer of sensitive visual patient data outside the clinical network.
Conversely, locally hosted models, though maintaining data sovereignty and offering better computational efficiency, lack the specialized, domain-specific knowledge critical for accurate surgical applications.
To bridge this gap, we introduce a training framework designed to yield a locally running and domain-optimized VLM for anatomical landmark identification and explanation in Complete Mesocolic Excision. \\
General-purpose LLMs like Gemini and ChatGPT possess extensive world knowledge, and when provided with sufficient context, can effectively mimic the responses of domain experts. 
To leverage this capability while maintaining data privacy, we propose feeding the general-purpose LLM with contextual information in text form, including the anatomy present and the current surgical phase. 
Additionally, we supply a binary mask detailing the local information of the anatomy. 
This method provides important context without directly leaking the original patient data.
This enables the automatic generation of high-quality, domain-specific training answers necessary for supervised fine-tuning (SFT) and subsequent refinement via Direct Preference Optimization (DPO) \cite{rafailov2023direct}.
The data generation process is partially supervised by a surgical expert to ensure factual correctness.
This approach facilitates the knowledge distillation from large, general-purpose LLMs into a smaller, private and domain-specific VLM suitable for secure, local deployment. \\
A rigorous evaluation demonstrates that our finetuned models provide explanations with higher factual and clinical accuracy compared to its base VLM.
Overall, this work presents a novel training framework for developing effective tools for surgeons using vision language models, using CME as an example.

\section{Methods}

\subsection{Dataset Generation}
We collected video data from 35 robot-assisted surgical procedures involving, among others, partial sigmoid or rectum colon resections. From these recordings, a senior surgical expert identified and isolated the video segments corresponding to the Complete Mesocolic Excision (CME) phase. A doctoral medical student performed semantic segmentation on individual frames extracted from the CME phases. The annotations targeted five anatomical structures important for intraoperative orientation during CME: preparation plane (1167), angel's hair (1009), vena mesenterica inferior (999), duodenum (184), and pancreas (414). The complete set of segmentations was subsequently reviewed and, where necessary, refined by two independent senior surgeons.

\subsection{VLM Training Framework}

The instruction-following training dataset for Supervised Fine-Tuning (SFT) of the Vision-Language Model (VLM), as outlined in Figure \ref{fig:Figure_1} (Step 1), comprises of a visual context, a textual prompt, and the expert answer.
The visual context is a composite image created by overlaying a semi-transparent colored segmentation mask onto the original endoscopic frame, intended to ground the VLM's visual attention.
The textual prompts are based on templates, such as "Identify the highlighted anatomical structure and explain its significance for Complete Mesocolic Excision".
The expert answer is generated by utilizing the semantic segmentation annotation to extract the structure's name and its precise location as a binary mask. 
This non-image information is used to prompt Gemini 2.5 Flash \cite{comanici2025gemini} to synthesize an expert-level, instruction-following response.
The generated responses are partially supervised by a surgical expert to ensure clinical correctness.
Importantly, using the binary mask and text ensures no actual intraoperative image data is transferred to the externally hosted LLM. \\
To further align the VLM to human preferences, Direct Preference Optimization (DPO) has emerged as a useful method \cite{rafailov2023direct}.
DPO directly optimizes the model's policy to maximize the likelihood of a preferred (accepted) answer over a non-preferred (rejected) answer, removing the need for an explicit reward model. 
Figure \ref{fig:Figure_1} (Step 2) illustrates the creation of the preference dataset utilized to instill the surgical expert preferences.
To ensure the training data is in-distribution for the DPO algorithm, we first utilize the Supervised Fine-Tuned VLM (from Step 1) with frozen weights. 
For each composite image and corresponding prompt, the SFT VLM generates five distinct answers by varying sampling parameters.
The five distinct generated answers and the ground-truth answer from Step 1 are used to prompt Gemini 2.5 Flash, to designate one accepted and one rejected responses based on clinical accuracy and relevance.
The resulting preference alignment dataset is then partially reviewed by a surgical expert to ensure clinical correctness. \\
Afterwards, the VLM model is trained with DPO using the created preference dataset as visualized in Figure \ref{fig:Figure_1} (Step 3).

\section{Experiments and Results}

\subsection{Implementation Details}
As our base VLM, we utilize the Qwen 2.5 VL 8B model \cite{Qwen2.5-VL}.
We conduct training on a NVIDIA H100 GPU and evaluate on a consumer RTX 4090.
The visual input is resized to a resolution of $960 \times 540$, with the attention-grounding visual context created by overlaying a semi-transparent green mask ($\alpha=0.5$) derived from the segmentation on the endoscopic frame.
The instruction-following dataset comprised $3156$ training and $790$ test image/conversation pairs.
For SFT, the model was trained for $1$ epoch using the AdamW optimizer with a learning rate of $1 \times 10^{-5}$ and an accumulated batch size of $8$. 
During SFT, the vision encoder was frozen, while the vision merger and language model were fully trained.
The subsequent DPO was conducted with a learning rate of $1 \times 10^{-6}$ and a preference-scaling factor of $\beta=0.1$. 
For DPO, all model components (vision encoder, visual merger, and language model) are fine-tuned.

\subsection{Evaluation}
We compare the performance of the base VLM to our SFT VLM and DPO VLM using automated metrics and a surgical expert evaluation on a held-out test set.
As automated metrics we use the BLEU-4 score and the BERTScore, which compare the n-gram overlap and the cosine similarity of BERT embeddings between the generated and ground-truth answers, respectively. 
We employ Gemini 2.5 Flash as an LLM-as-a-Judge, leveraging its ability to provide a scalable and explainable approximation of human preferences in evaluation \cite{zheng2023judging}.
Additionally, a surgical expert evaluation is conducted on $40$ randomly selected image/prompt pairs, providing a direct assessment of clinical utility.
Both the LLM Judge as well as the expert evaluation is based on a weighted  composite score consisting of clinical/factual accuracy ($75\%$), which evaluates the correctness of anatomical identification, position, and CME significance, and conciseness ($25\%$) assessing avoidance of repetition.
In addition, for each of the 40 image/prompt pairs, the surgical expert selects the model with the preferred answer in terms of clinical correctness. Multiple preferred models can also be selected for each image/prompt pair.
\begin{table}[t]
\centering
\caption{Automated and expert evaluation results for Base, SFT, and DPO VLMs. 
BLEU and BERTScore range from 0–1, while Gemini and expert scores range from 1–5.}
\label{tab:compact_eval}
\resizebox{\columnwidth}{!}{%
\begin{tabular}{lccc}
\toprule
\textbf{Metric} & \textbf{Base VLM} & \textbf{SFT VLM} & \textbf{DPO VLM} \\
\midrule
\multicolumn{4}{l}{\textbf{Automated Evaluation}} \\
\midrule
BLEU ↑ (0–1) & 0.11 ± 0.04 & 0.28 ± 0.09 & \textbf{0.27 ± 0.09} \\
BERTScore ↑ (0–1) & 0.26 ± 0.04 & \textbf{0.55 ± 0.20} & 0.54 ± 0.10 \\
LLM Judge (Gemini) ↑ (1–5) & 1.76 ± 0.27 & 3.85 ± 1.04 & \textbf{3.91 ± 1.09} \\
\midrule
\multicolumn{4}{l}{\textbf{Expert Evaluation (n = 40 cases)}} \\
\cmidrule(lr){1-4}
Expert score ↑ (range 1–5) & 1.05 ± 0.1 & \textbf{4.50 ± 0.57} & 4.45 ± 0.58 \\
Preferred model ↑ (count) & 0 & 33 & 26 \\
\bottomrule
\end{tabular}%
}
\end{table}
\subsection{Results}
\begin{figure}[t]
\centering
\centerline{\includegraphics[width=\columnwidth]{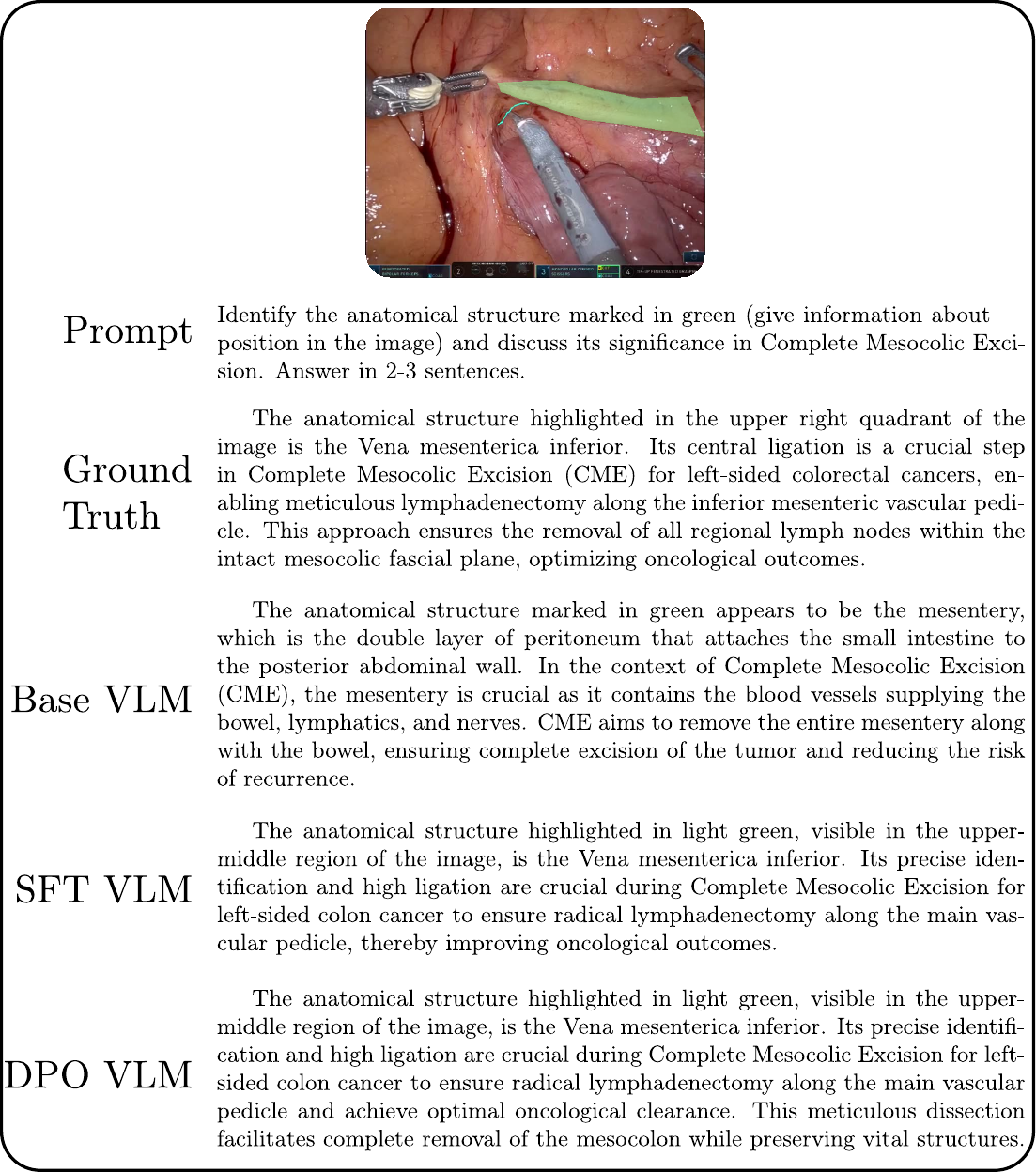}}
\caption{Qualitative results of all utilized VLMs in this work. Each model is given the input image with the highlighted structure as well as the Prompt.}\medskip
\label{Fig.2}
\end{figure}
As shown in Table \ref{tab:compact_eval}, the Base VLM shows significant disadvantages in terms of correct answers, reflected by the lowest scores across all metrics (e.g. LLM Judge: $1.76 \pm 0.27$).
Qualitatively, its responses consistently collapsed to identifying only "mesentery," irrespective of the visually highlighted structure (e.g., Figure 2).
Supervised Fine-Tuning (SFT) using the instruction-following dataset results in the largest performance increase.
The SFT VLM achieves an increase in both semantic understanding (BERTScore: $0.55 \pm 0.20$) and clinical quality (LLM Judge: $3.85 \pm 1.04$).
The subsequent preference alignment via DPO results in a marginal, non-significant increase in overall performance compared to SFT.
Analyzing the performance by anatomical structure shows that the DPO VLM achieves the highest LLM Judge scores for preparation plane ($4.0$) and duodenum ($4.26$), while the lowest performance was observed for angel's hair ($3.5$). \\
The evaluation of the different models by the surgical expert shows high agreement with the automated metrics' findings.
While the base VLM provides almost exclusively incorrect clinical responses, the SFT and DPO models generally achieve higher absolute scores for clinical accuracy than the LLM Judge benchmark. 
Notably, the SFT and DPO models generate identical responses frequently.
Overall, the SFT model provides a more comprehensive clinical response in several instances and is consequently selected as the preferred model more often.
However, both the SFT and DPO models produce instances of laterality error, whereby left-sided CME is incorrectly identified as right-sided, and vice versa.
\section{Discussion and Conclusion}
In this work, we introduce a privacy-preserving training framework to create a locally running, efficient, and CME domain-optimized VLM.
We leverage the ability of general-purpose LLMs to mimic clinicians' responses while ensuring data privacy by feeding the general-purpose LLM with contextual information such as anatomy name, current surgical phase, and binary segmentation mask of the anatomical position.
We conduct a comprehensive evaluation using both automated evaluation metrics and expert evaluation based on a smaller test set.
The results confirm that the Base VLM can not transfer general knowledge to the specific surgical task.
Our semi-automatic data synthesis approach for SFT yields the most significant performance gain, demonstrating that this method can successfully inject surgical domain knowledge.
The subsequent Direct Preference Optimization (DPO) offered no improvements, evidenced by the surgical expert's evaluation. 
We hypothesize that this can be a confirmation of the high quality alignment already present in our expert-supervised SFT dataset, which left minimal error for DPO to correct. \\
To guide the VLM for anatomy explanation, we use visual cues, i.e. transparent overlayed mask, in this work. To enable an end-to-end system, the integration of automatic segmentation should be further considered.
Furthermore, in the case of domain-specific VLMs, it is of interest to improve and evaluate the deeper contextual knowledge of these models, for example using multi-turn, conversational data. This enables the development of simple Q$\&$A tools into a dynamic surgical tutor. \\
In conclusion, this work demonstrates a data-efficient and privacy-conforming pathway to distill knowledge from large, cloud-based models into secure, local VLMs. Our framework successfully produces a specialized model capable of accurately recognizing and explaining important anatomical structures in CME, validating its potential for surgical education and intraoperative support. \\
\hrulefill \\
\textbf{Ethical approval} The utilized data was fully anonymized and not made publicly available. \\
\textbf{Funding} This publication/work was partially funded/co-funded by the European Union under Horizon Europe programme grant agreement No. 101059903.

\bibliographystyle{IEEEbib}
\bibliography{strings,refs}

\end{document}